%% file: main.tex
\documentclass[11pt,a4paper]{article}
\usepackage[hyperref]{acl2020}
\usepackage{times}
\usepackage{amsmath}
\usepackage{multirow}
\usepackage{latexsym}
\usepackage{graphicx, caption, subcaption}
\usepackage{todonotes}
\usepackage{enumitem,kantlipsum}

\usepackage{microtype}

\aclfinalcopy 

\title{Echo State Neural Machine Translation}

\author{Ankush Garg \quad Yuan Cao \quad Qi Ge \\
  Google Research \\
  \texttt{\{ankugarg, yuancao, qge\}@google.com} \\}

\date{}

\begin{document}
\maketitle

\input{abstract}

\input{introduction}

\input{background}

\input{model}

\input{experiments}

\input{related_work}

\input{conclusion}

\bibliography{anthology,acl2020}
\bibliographystyle{acl_natbib}

\end{document}

%% file: abstract.tex

\begin{abstract}
We present neural machine translation (NMT) models inspired by echo state network (ESN), named Echo State NMT (ESNMT), in which the encoder and decoder layer weights are randomly generated then fixed throughout training. We show that even with this extremely simple model construction and training procedure, ESNMT can already reach 70-80\% quality of fully trainable baselines. We examine how spectral radius of the reservoir, a key quantity that characterizes the model, determines the model behavior. Our findings indicate that randomized networks can work well even for complicated sequence-to-sequence prediction NLP tasks.
\end{abstract}

%% file: introduction.tex
\section{Introduction}

In this paper we propose a neural machine translation (NMT) model in which the encoder and decoder layers are randomly generated then fixed throughout training. We show that despite the extreme simplicity in model construction and training procedure, the model still performs surprisingly well, reaching 70-80\% BLEU scores given by a fully trainable model of the same architecture.

Our proposal is inspired by Echo State Network (ESN) \citep{jaeger2001,maass2002}, a special type of recurrent neural network (RNN) whose recurrent and input matrices are randomly generated and untrained. Such a model building procedure is counter-intuitive, however as long as its dynamical behavior (characterized by a few key model hyperparameters) properly approximates the underlying dynamics of a given sequence processing task, randomized models can also yield competitive performance. If we view language processing from a dynamical system's perspective \citep{Elman1996}, ESN can be an effective model for NLP tasks as well.

There are existing works that apply randomized approaches similar to ESN to NLP tasks \citep{tong2007,hinaut2012,wieting2018,enguehard2019}, which report the effectiveness of using representations produced by random encoders. However the capability of ESN in directly handling more general and complicated sequence-to-sequence (seq2seq) prediction tasks has not been investigated yet. 


\paragraph{Contribution} We propose an Echo State NMT model with a randomized encoder and decoder, extending ESN to a challenging seq2seq prediction task, and study its uncharacteristic effectiveness in MT. This also provides an interesting opportunity for model compression, as one only needs to store one single random seed offline, from which all randomized model components can be deterministically recovered.

%% file: background.tex
\section{Background} \label{sec:background}
Echo State Network \citep{jaeger2001} is a special type of recurrent neural network, in which the recurrent matrix (known as ``reservoir'') and input transformation are randomly generated then fixed, and the only trainable component is the output layer (known as ``readout''). A very similar model named Liquid State Machine (LSM) \citep{maass2002} was independently proposed almost simultaneously, but with a stronger focus on computational neuroscience. This family of models started by ESN and LSM later became known as Reservoir Computing (RC) \citep{Verstraeten2007}.

A basic version of ESN has the following formulation :
\begin{align}
    \mathbf{h}_{t} &= \text{tanh} \left(\mathbf{W}_{res}\mathbf{h}_{t-1} + \mathbf{W}_{in} \mathbf{x}_t\right) \label{eq:esn} \\
    \mathbf{y}_t &= f(\mathbf{W}_{out}\mathbf{h}_{t}) \notag
\end{align}
in which $\mathbf{h}_t$ and $\mathbf{x}_t$ are the hidden state and input at time $t$,  $\mathbf{y}_t$ is the output and $f$ being a prediction function (for example softmax for classification). This formulation is almost equivalent to a simple RNN, except that the reservoir and input transformation matrices $\mathbf{W}_{res}$ and $\mathbf{W}_{in}$ are randomly generated and fixed. $\mathbf{W}_{res}$ is also often required to be a sparse matrix. The only component that remains to be trained is the readout weights $\mathbf{W}_{out}$. 

Despite the extremely simple construction process of ESN, it has been shown to perform surprisingly well in many regression and time-series prediction problems. A key condition for ESN to function properly is called the Echo State Property (ESP) \citep{jaeger2001,Yildiz2012}, which basically claims that the ESN states asymptotically depend only on the driving input signals (hence states are ``echos'' of inputs), while the influence of the initial states vanishes over time. ESP essentially requires the recurrent network to have a ``fading memory'', which is also shown to be critical in optimizing a dynamical system's computational capacity \citep{Legenstein2007,Dambre2012}.

Theoretical analysis shows that in order for ESP to hold, the spectral radius of the reservoir matrix $\rho(\mathbf{W}_{res})$, defined as the largest absolute value of its eigenvalues, needs to be smaller than 1, although it was argued that this is not a rigorous condition \citep{Pascanu2011}. Intuitively, $\rho(\mathbf{W}_{res})$ determines how long an input signal can be retained in memory: smaller radius results in a shorter memory while larger radius enables a longer memory. In addition, the scale of the input, which determines how strong inputs influence the dynamics, remains a hyperparameter critical to the performance of the model.

Recently ESNs have also been extended to deep versions in which multiple recurrent layers are stacked up \citep{Gallicchio2017,Gallicchio2017b,Gallicchio2017c,Gallicchio2019a,Gallicchio2019}. It has been shown that different levels of the ESN layers are able to capture signal dynamics at different scales. 




%% file: model.tex
\section{Echo State Neural Machine Translation}
\subsection{Model Architecture}\label{sec:model_arch}
Inspired by the simple yet effective construction of ESN, we are interested in extending ESN to challenging sequence-to-sequence prediction tasks, especially NMT. We propose an ESN-based NMT model whose architecture follows RNMT+ \cite{chen2018}, the state-of-the-art RNN-based NMT model. Unlike RNMT+ which is fully trainable, we simply replace all recurrent layers in the encoder and decoder with echo state layers as shown in Eq.~\ref{eq:esn}, and call this model ESNMT.

In addition to the simple RNN cell employed by the original ESN (Eq.~\ref{eq:esn}), we also explore a variation of ESNMT which employs the LSTM cell \citep{Hochreiter1997}. That is, we randomly generate all weight matrices in the LSTM and keep them fixed. We call this version ESNMT-LSTM.


In the models above, the trainable components are word embedding, softmax and attention layers. Instead of freezing both encoder and decoder, we also investigate settings where only the encoder or decoder is frozen. We further consider cases where even the attention and embedding layers are randomized and fixed. These variations of architectures are compared in Sec.\ref{sec:exp_ablation}.

We note that the size of the reservoir can be cheaply increased since they do not need to be trained, which often leads to better performance. We nevertheless constrain the ESNMT model size to be the same as trainable baselines in our experiments, even though the latter contain way more trainable parameters.

\subsection{Adaptive Echo State Layers}
As described in Sec.\ref{sec:background}, two critical hyperparameters that determine the dynamics of ESN and its performance are the spectral norm of the reservoir matrix and input scale. While common practice manually tunes these hyperparameters for specific tasks, we treat them as trainable parameters and let the training procedure find suitable values. Specifically, we modify the ESN layer in Eq.~\ref{eq:esn} into

\begin{align}
    \mathbf{h}_t^l = \text{tanh} \left(\rho^l \mathbf{W}_{res}^{l}\mathbf{h}_{t-1} + \gamma^l \mathbf{W}_{in}^{l} \mathbf{x}_t\right) \label{eq:esn_new}
\end{align}

\noindent where $\rho^l$ and $\gamma^l$ are learnable scaling factors for the reservoir of the $l^{th}$ layer and input transformation matrices respectively. Similar modification is applied to the LSTM state transition formulation in ESNMT-LSTM.


\subsection{Training}
Our models are trained with back-propagation and cross-entropy loss as usual.\footnote{Note that ESNs have been commonly applied to regression and time-series prediction problems, in which case the loss functions are usually mean square error and readout weights can be updated in close-form without the necessity of back-propagation.} Note that since recurrent layer weights are fixed and their gradients are not calculated, the challenging gradient explosion/vanishing problem \citep{Pascanu2013} commonly observed in training RNNs can be significantly alleviated. Therefore we expect no significant difference in quality between ESNMT and ESNMT-LSTM, since the LSTM architecture, which was originally designed to tackle the gradient instability problem, will not be superior in this case. This is verified in our experimental results (Sec. \ref{sec:exp_main}).

\subsection{Model Compression}
Since randomized components of ESNMT can be deterministically generated simply from one fixed random seed, to store the model offline we only need to save this single seed together with remaining trainable model parameters. For example, in an ESNMT-LSTM model with 6-layer encoder and decoder of dimension 512 and vocabulary size 32K, around $52\%$ of the parameters from the recurrent layers can be recovered from a single random seed.

%% file: experiments.tex
\section{Experiments}\label{sec:exp}
\subsection{Setup}
We train and evaluate our models on WMT'14 English$\rightarrow$French, English$\rightarrow$German and WMT'16 English$\rightarrow$Romanian datasets. Sentences are processed into sequences of sub-word units using BPE \citep{sennrich2016}. We use a shared vocabulary of 32K sub-word units for both source and target languages.

Our baselines are fully trainable RNMT+ with LSTM cells. For the proposed ESNMT models, all reservoir and input transformation matrices are generated randomly from a uniform distribution between -1 and 1. The reservoirs are then randomly pruned so that $\mathbf{W}_{res}$ and $\mathbf{W}_{in}$ reach 20-25\% sparsity\footnote{We also experimented with other sparsity levels, but did not observe significant differences in model quality.}, and normalize $\mathbf{W}_{res}$ so that its spectral radius equals to 1. Note the effective spectral radius and input scaling are determined by the learnable scaling factors as shown in Eq.~\ref{eq:esn_new}, which are initialized to 1 and 10 respectively for all layers. For all models the number of encoder and decoder layers are equally set to 6, and model dimension to 512 or 2048. We also adopt similar training recipes as used by the RNMT+ \citep{chen2018}, including dropout, label smoothing and weight decay for all our models.

\subsection{Results} \label{sec:exp_main}
Table~\ref{tbl:main} compares BLEU scores for all languages pairs given by different models. 

\begin{table}[htbp]
\small
\centering
\begin{tabular}{lclll}
\hline \textbf{Model} &\textbf{Dim} &\textbf{EnFr} & \textbf{EnDe} & \textbf{EnRo} \\ \hline
\multirow{2}{*}{LSTM} &512 & 39.15 & 26.75 & 23.82 \\
                      &2048 & 40.33 & 27.42 & 24.69 \\
\hline
\multirow{2}{*}{ESNMT} &512 & 31.04 &18.77 &20.05 \\
                       &2048 & 32.53 &19.16 &\textbf{21.01} \\
\multirow{2}{*}{ESNMT-LSTM} &512 & 31.15 &18.92 &20.01 \\
                       &2048 & \textbf{32.61} &\textbf{19.86} &20.90 \\
\hline
\end{tabular}
\caption{\label{tbl:main} BLEU score comparison on WMT testsets. LSTM is the fully-trainable baseline following RNMT+ settings, ESNMT(-LSTM) are proposed models with simple RNN or LSTM cells. Best ESNMT numbers are in bold.}
\end{table}



The results show that ESNMT can reach 70-80\% of the BLEU scores yielded by fully trainable baselines across all settings. Moreover, using LSTM cells yields more or less the same performance as a simple RNN cell. This verifies our hypothesis in Sec.~\ref{sec:model_arch} that an LSTM cell is not particularly advantageous compared to a simple RNN cell in the ESN setting.



\paragraph{Ablation study} \label{sec:exp_ablation}
As mentioned in Sec~\ref{sec:model_arch}, in addition to randomizing both the encoder and decoder, we explore other strategies of applying randomization, and conduct an ablation test as follows: We start by randomizing and freezing everything in the ESNMT-LSTM model (dimension 512) except the softmax layer, then gradually release attention, encoder and/or decoder so that they become trainable. The results for En$\rightarrow$Fr are shown in Table~\ref{tbl:ablation_study}.

\begin{table}[htbp]
\small
\centering
\begin{tabular}{llll}
\hline \textbf{Model} &\textbf{BLEU} \\ \hline
ESNMT-LSTM-512 (softmax only) & 4.44 \\
+ Embedding &26.63 \\
+ Attention &31.15 \\
+ Encoder only &37.98 \\
+ Decoder only &35.21 \\
+ Encoder \& decoder (fully trainable) &39.15 \\
\hline
\end{tabular}
\caption{\label{tbl:ablation_study} First row corresponds to ESNMT-LSTM of dimension 512 in which \emph{only softmax} is trainable. Next few rows gradually make embedding, attention, encoder \emph{or} decoder layers trainable. Last row corresponds to the fully trainable LSTM baseline.}
\end{table}

From the table we have the following interesting findings:
\begin{enumerate}[topsep=0pt, itemsep=0mm, wide, labelwidth=!, labelindent=0pt]
    \item By randomizing \emph{only} the entire decoder, the BLEU score (37.98) drops only by 1.17 from the baseline (39.15).
    \item Randomizing the encoder incurs more BLEU loss (35.21) than decoder. This shows that training the encoder properly is more critical to seq2seq tasks.
    \item Embedding layer deserves the most training. It lifts the BLEU given by an almost purely randomized model (4.44) immediately to 26.63. It is also interesting to note that a model with only the embedding and softmax layers trainable is already able to reach this BLEU score.
\end{enumerate}

\begin{figure}
\centering
\includegraphics[width=\linewidth]{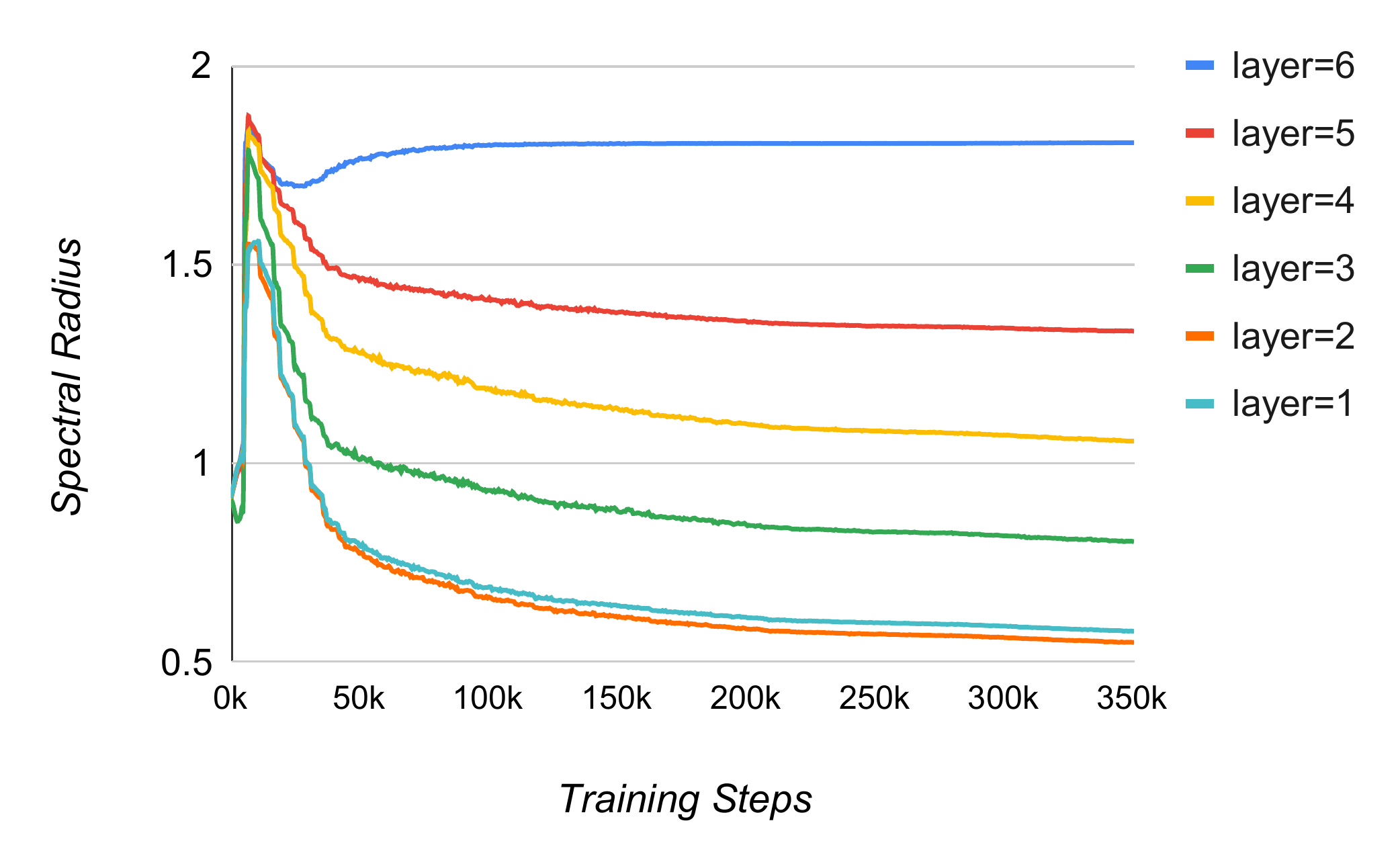}
\caption{Learning curves of spectral radius $\rho^l$ for each forward encoder ESN layer.}
\label{fig:enc_radius}
\end{figure}

\paragraph{Effect of spectral radius}
To find out why ESNMT works, we examine learned spectral radii $\rho^l$ for each layer, which are are critical in characterizing the dynamics of ESNMT. In Fig.~\ref{fig:enc_radius} we show the learning curves of $\rho^l$ for all layers in the forward encoder ESN. The figure shows a clear trend that the radius increases almost monotonically from bottom to top layer (0.55 to 1.8). This indicates that lower layers retain short memories and focus more on word-level representations, while upper layers keep longer memories and account for better sentence-level semantics which requires capturing long-term dependencies between inputs. Similar phenomena are observed for the backward encoder ESN and decoder.

\begin{figure}
\centering
\includegraphics[width=\linewidth]{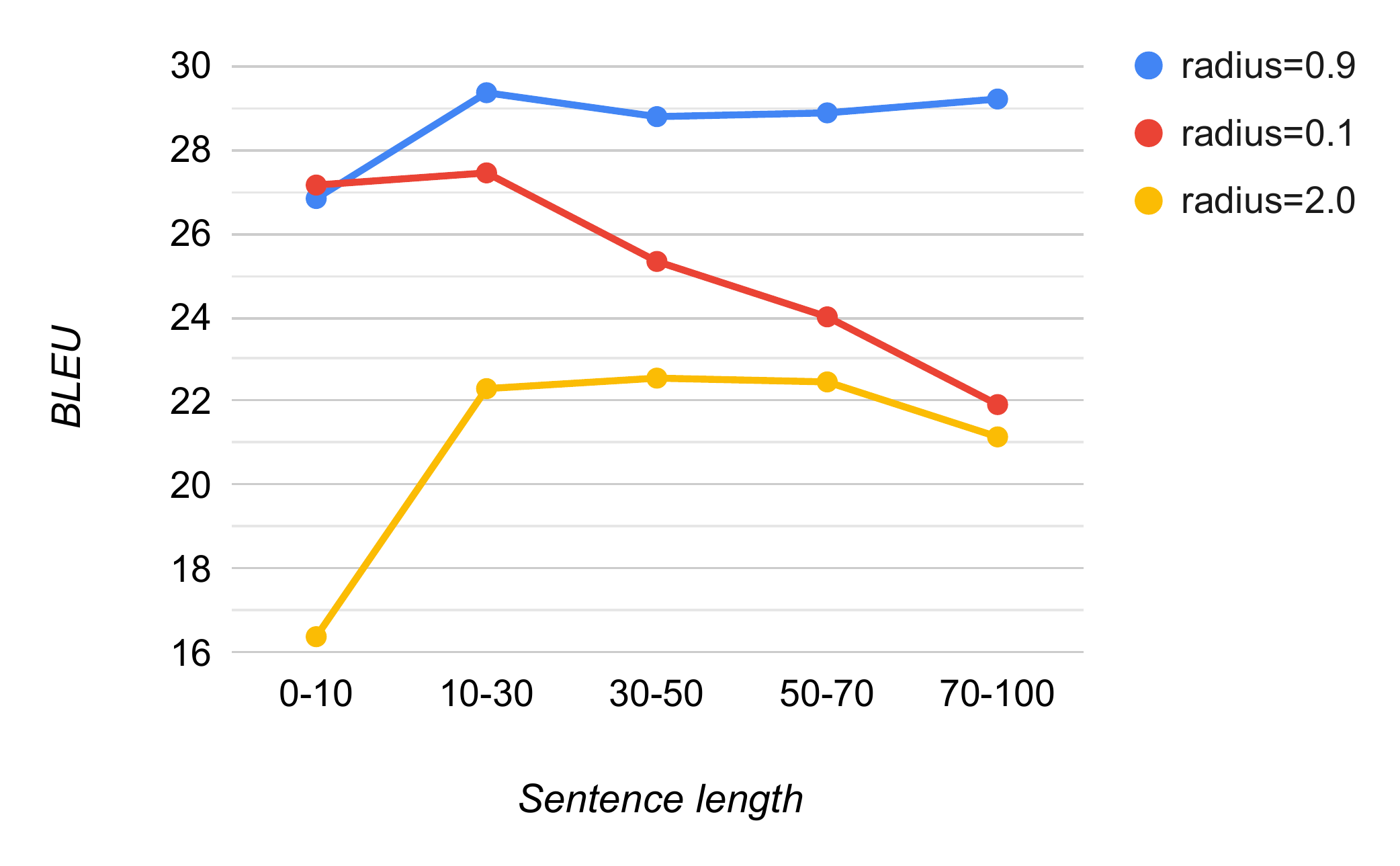}
\caption{Sentence lengths vs. BLEU on EnFr testset, with different fixed spectral radius values.}
\label{fig:length_vs_radius}
\end{figure}

To further investigate how spectral radius determines translation quality, we study BLEU scores on EnFr testset as a function of sentence length, using models in which radii $\rho^l$ are fixed for all layers and set to 0.1, 0.9 or 2.0. The results are shown in Fig.~\ref{fig:length_vs_radius}, from which we see that when the radius is small (0.1), the model favors shorter sentences which requires less memory, increasing the radius to 2.0 equips the model with non-fading memory, in which remote inputs outweigh recent inputs, resulting in worse quality on short sentences. Radius 0.9 maintains a good balance between short and long memories, yielding the best quality. Nevertheless the overall quality for all settings is worse than models whose radii are learned (Table~\ref{tbl:main}).

%% file: related_work.tex
\section{Related Work}
Perhaps most related to our work are \citep{wieting2018,enguehard2019}, in which they studied the effectiveness of using randomized encoders in performing SentEval tasks. We also note that similar randomization approaches have also been applied to other NLP problem settings \citep{tong2007,hinaut2012,yukinori2013,alhama2016,zhang2018,tenney2018,Ramamurthy2019}. However none of them explore the potential of ESN for encoder-decoder models or more challenging tasks like MT, nor did they study why randomization in these problem settings works properly. These are the questions we aim to address in our paper.


%% file: conclusion.tex
\section{Conclusion}
We proposed Echo State NMT models whose encoder and decoder are composed of randomized and fixed ESN layers. Even without training these major components, the model can already reach 70-80\% performance yielded by fully trainable baselines. These surprising findings encourage us to rethink about the nature of encoding and decoding in NMT, and design potentially more economic model architectures and training procedures.

ESNMT is based on the recurrent network architecture. One interesting research problem for future exploration is how to apply randomized algorithms to non-recurrent architectures like Transformer \citep{ashish2017}. This is potentially possible, as exemplified by randomized feedforward networks like Extreme Learning Machine \citep{Huang2006}.